\documentclass{article}

\usepackage{PRIMEarxiv}

\usepackage[utf8]{inputenc} 
\usepackage[T1]{fontenc}    
\usepackage{hyperref}       
\usepackage{url}            
\usepackage{booktabs}       
\usepackage{amsfonts}       
\usepackage{nicefrac}       
\usepackage{microtype}      
\usepackage{lipsum}
\usepackage{fancyhdr}       
\usepackage{graphicx}       
\graphicspath{{media/}}     

\usepackage{authblk} 
  
\usepackage{array}
\usepackage{subcaption}
\usepackage{amsmath}
\usepackage{multirow}

\usepackage{xspace}
\newcommand{\ourmethod}{{DATS}\xspace}
  
\title{Local vs. Global: Local Land-Use and Land-Cover Models Deliver Higher Quality Maps
}
\author{%
    Girmaw Abebe Tadesse\textsuperscript{1}\thanks{\textsuperscript{}Corresponding author: \textbf{gtadesse@microsoft.com}},\hspace{0.2cm}    
    Caleb Robinson\textsuperscript{1},\hspace{0.2cm}    
    Charles Mwangi\textsuperscript{2},\hspace{0.2cm}    
    Esther Maina\textsuperscript{2},\hspace{0.2cm}    
    Joshua Nyakundi\textsuperscript{2},\hspace{0.2cm} \\ 
    Luana Marotti\textsuperscript{1},\hspace{0.2cm}    
    Gilles Quentin Hacheme\textsuperscript{1},\hspace{0.2cm}    
    Hamed Alemohammad\textsuperscript{3},\hspace{0.2cm}    
    Rahul Dodhia\textsuperscript{1},\hspace{0.2cm} \\   
    Juan M. Lavista Ferres\textsuperscript{1}  
}  
  
\affil{\textsuperscript{1}\footnotesize\textit{Microsoft AI for Good Research Lab}\hspace{0.5cm}\textsuperscript{2}\footnotesize\textit{Kenya Space Agency}\hspace{0.5cm}\textsuperscript{3}\footnotesize\textit{Clark University}}    

\begin{document}
\maketitle

\begin{abstract}
In 2023, $58.0\%$ of the African population experienced moderate to severe food insecurity, with $21.6\%$ facing severe food insecurity. Land-use and land-cover maps provide crucial insights for addressing food insecurity by improving agricultural efforts, including mapping and monitoring crop types and estimating yield. The development of global land-cover maps has been facilitated by the increasing availability of earth observation data and advancements in geospatial machine learning. However, these global maps exhibit lower accuracy and inconsistencies in Africa, partly due to the lack of representative training data.
To address this issue, we propose a data-centric framework with a teacher-student model setup, which uses diverse data sources of satellite images and label examples to produce local land-cover maps. Our method trains a high-resolution teacher model on images with a resolution of $0.331$ $\mathtt{m/pixel}$ and a low-resolution student model on publicly available images with a resolution of $10$ $\mathtt{m/pixel}$. The student model also utilizes the teacher model's output as its weak label examples through knowledge transfer.
We evaluated our framework using Murang'a county in Kenya, renowned for its agricultural productivity, as a use case. Our local models achieved higher quality maps, with improvements of $0.14$ in the $F_1$ score and $0.21$ in Intersection-over-Union, compared to the best global model. Our evaluation also revealed inconsistencies in existing global maps, with a maximum agreement rate of $0.30$ among themselves.
Our work provides valuable guidance to decision-makers for driving informed decisions to enhance food security.
\end{abstract}


\section{Introduction}\label{sec:introduction}

Land-Use and Land-Cover (LULC) maps are critical for monitoring 14 of the 17 United Nations Sustainable Development Goals (SDGs)~\cite{alemohammad2020landcovernet}. LULC maps enable informed resource management, urban planning, environment monitoring to enhance food security~\cite{kerner2024accurate}. 
In Africa, $58.0\%$ of the population experienced moderate to severe food insecurity in 2023, while $21.6\%$ faced severe food insecurity, suggesting that by 2030, $53\%$ of the global population facing hunger will be concentrated in the continent~\cite{fao2024state}.
In addition, most of the economies in Sub-Saharan Africa (SSA) are dependent on the agriculture sector~\cite{hacheme2024weak,diao2010role}. For example, in Kenya, the sector generates $60\%$ of foreign exchange, accounts for $70\%$ of employment, produces approximately $45\%$ of total government revenue, and supplies $75\%$ of raw materials for industry~\cite{wanzala2024impact}. However, the sector faces multiple challenges, such as unpredictable weather, soil degradation, competing land use,
and inadequate agricultural extension service, often exacerbated by adverse climate impacts and expansion in population settlements - resulting in a growing prevalence of food insecurity~\cite{giller2020food}. 

Other challenges to the sector in the SSA include the European Union Anti-deforestation Law (EUAL) that aims to block agricultural products grown on deforested lands from accessing European markets~\cite{muradian2025will}. EUAL poses a risk for small scale farmers that are limited in technological resources but still are the leading agricultural producers. For example, $80\%$ 
of the coffee consumed worldwide is produced by smallholder farmers~\cite{eual}, and seven out of the top ten global coffee markets for Kenya are in the European economic zone~\cite{foodbusinessafrica}.

LULC maps play a vital role in supporting the agriculture sector by first characterizing land uses and land covers, such as croplands, forests, and water bodies, and subsequently automating downstream tasks like crop mapping, monitoring, and yield estimation~\cite{brown2022dynamic,kerner2024accurate}. The growing availability of remotely sensed data, along with geospatial machine learning models, helps to build global LULC maps~\cite{rolf2024mission}. These include Google's Dynamic World (GDW)~\cite{brown2022dynamic}, European Space Agency's (ESA) WorldCover~\cite{zanaga2022esa}, and Environmental Systems Research Institute's (ESRI) LULC~\cite{karra2021global}. However, there is still a significant disparity in LULC mapping efforts in Africa when compared to Europe and North America~\cite{alemohammad2020landcovernet,venter2022global,xu2019comparisons,nabil2020assessing}. In addition, the global maps are reported to exhibit lower accuracy and inconsistencies in Africa~\cite{kerner2024accurate,rolf2022striving}, partly due to the lack of quality data representing the region~\cite{alemohammad2020landcovernet} or failure of global models to capture meaningful variation within sub-regions~\cite{rolf2022striving}. The performance and trustworthiness of ML models depend critically on the quality of data~\cite{oala2023dmlr,rolf2022striving, liang2022advances}.
While satellite images with a sub-meter resolution offer high quality details to achieve accurate LULC maps, their accessibility is often limited and expensive compared to lower-resolution data. The collection of lower-resolution products is typically scheduled and can be used in various applications. However, the collection of high-resolution products is task-driven and usually time-sensitive. Recently, there has been a growing interest to use both high-and lower-resolution data, e.g., using a teacher-student model setup~\cite{sirko2023high}. A teacher-student model is a machine learning framework where a larger or more complex model, also known as the ``teacher'' model, is used to train a smaller or simpler ``student'' model~\cite{wang2021knowledge}.

\begin{figure}[t]
    \centering
    \includegraphics[width=0.99\linewidth]{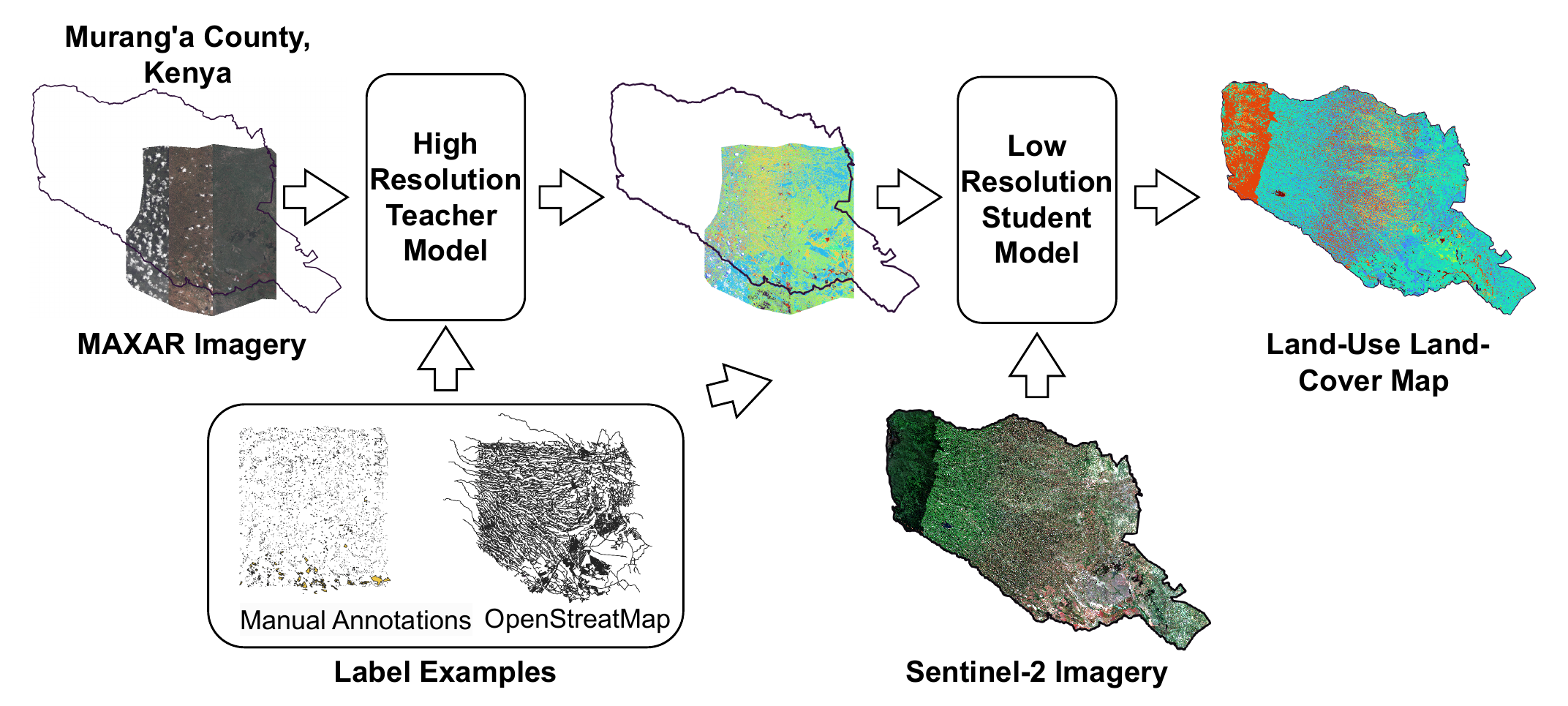}
    \caption{\textbf{Overview of our framework to build local land-use and land-cover (LULC) model that produces high quality map, using Murang'a county in Kenya as our area of interest}. We propose a setup of teacher and student models to be trained on high- and low-resolution satellite images, respectively. }\label{fig:overview}
\end{figure}

 To this end, we propose \textbf{DA}ta-centric framework with a \textbf{T}eacher-student model \textbf{S}etup (\ourmethod), which uses diverse data sources of satellite images and label examples to produce high-quality and local LULC maps (see Fig.~\ref{fig:overview}). The models are implemented with deep learning-based semantic segmentation architectures. 
We selected Murang'a county in Kenya, renowned for its agricultural productivity, as our Area of Interest (AOI). The county is situated between latitudes \(0^\circ 34' \, \text{S}\) and \(1^\circ 07' \, \text{S}\), and longitudes \(36^\circ 00' \, \text{E}\) and \(37^\circ 27' \, \text{E}\). The county spans a total area of approximately \(2,558.8 \, \text{km}^2\).
Here are the specific contributions of this work:
\begin{enumerate}
\item We propose a framework with a teacher-student model setup to build a local LULC map that uses diverse datain image resolutions and sources of label examples. Specifically, we train the teacher model using a high-resolution Maxar imagery ($0.331$ $\mathrm{m/pixel}$) with limited spatial coverage, which produces a high-resolution LULC map for a portion of our AOI. We then train the student model using lower-resolution Sentinel-2 imagery ($10$ $\mathrm{m/pixel}$) and the predictions from the teacher model as weak labels in the form of knowledge transfer~\cite{wang2021knowledge}. 

\item We evaluate the performance of our framework using different validation sets in a close collaboration with domain experts. 
 Our approach has resulted in the production of a high-quality LULC map that covers Murang'a county in Kenya. This map is now being used for downstream agricultural tasks such as crop type mapping and yield estimation by partner organizations. 

\item We compare the maps generated by our local model with those generated by multiple existing global models. Our analysis highlights the limitations of the global models in Africa, including lower accuracy and inconsistencies. Our model achieves improvements of 0.14 in the $F_1$ score and 0.21 in Intersection-over-Union, compared to the best global model. 
\end{enumerate}

\section{Data}\label{sec:data}
Our framework, shown in Fig.~\ref{fig:overview}, uses raw \textit{Satellite Imagery} and \textit{Label Examples} (annotations), collected from different sources, to train and test the models. 
\subsection{Satellite Imagery}\label{subsec:satellite_imagery}
Our teacher and student models use imagery sources with different resolutions. The teacher model uses a high-resolution Maxar imagery, with $0.331$ $\mathtt{m/pixel}$ resolution and red, green and blue channels, collected in 2022. However, it only covers $51.55\%$ of our AOI. In addition to its limited coverage, access to the high-resolution Maxar imagery is expensive which limits scalability. For example, Maxar's GeoEye-1 and WorldView-2-4 50 cm 3-Band satellite imagery products have a minimum cost of $\$17.50$ and $\$27.50$ per $km^2$, respectively, for archived and new imagery~\cite{landinfo}. To this end, we use a student model that takes publicly available multispectral Sentinel-2 images, with $10$ \textit{m/pixel}. We derived the median composite from all the Sentinel-2 images from 2022 for the entire AOI, including the red, green, blue, near infrared and short-wave infrared channels.

\begin{table}[t]  
    \centering  
        \caption{\textbf{The number (\#) of example polygons used for training and testing our High-Resolution Teacher Model and Low-Resolution Student Model} The teacher model was trained using Maxar imagery ($0.331$ $\mathtt{m/pixel}$ resolution), covering only a portion of Murang'a county - our area of interest (AOI), where as the student model was trained on publicly available but lower resolution Sentinel-2 imagery ($10$ $\mathtt{m/pixel}$ resolution), covering the entire AOI.}\label{table:combined_lulc}
    \resizebox{0.7\linewidth}{!}{%
    \begin{tabular}{ll|ll}  
        \toprule  
        \multicolumn{2}{c}{\textbf{Teacher Model}} & \multicolumn{2}{c}{\textbf{Student Model}} \\ 
       \textbf{LULC class} & \textbf{\# of polygons} & \textbf{LULC class} & \textbf{\# of polygons} \\  
        \midrule  
        Bare Ground & 1356 & Bare Ground & 187163 \\  
        Built-up & 2158 & Built-up & 276411 \\  
        Crop & 1814 & Crop & 483263 \\  
        Flooded Vegetation & 308 & Grass & 169395 \\  
        Grass & 1288 & Road & 256147 \\  
        Shrub \& Scrub & 1438 & Shrub \& Scrub & 289540 \\  
        Trees & 1796 & Trees & 235708 \\  
        Water & 1224 & Water & 28409 \\  
        \midrule  
        \textbf{Total} & \textbf{11382} & \textbf{Total} & \textbf{1892036} \\  
        \bottomrule  
    \end{tabular}  
    }    
\end{table}  
  
\subsection{Label Examples}\label{subsec:annotations}
The LULC classes in this work are adopted from~\cite{brown2022dynamic}: \textit{Bare Ground}, \textit{Built-up}, \textit{Crop}, \textit{Flooded Vegetation}, \textit{Grass}, \textit{Shrub \& Scrub}, \textit{Trees} and \textit{Water}.
Given the diverse nature of built-up areas, 
 we split the \textit{Built-up} class into separate classes of \textit{Building} and \textit{Road} during the training of the teacher model. 
We adopted two strategies to collect label examples for our models. First, domain experts were recruited to annotate label examples as polygons, primarily using the high-resolution Maxar imagery. We used Microsoft's satellite imagery labeling toolkit~\cite{microsoftlabeling} for our annotation efforts.
The domain experts annotated $11,382$ geographically diverse polygons, and the distribution of the polygons across the LULC classes is shown in Table~\ref{table:combined_lulc}. 
Second, we used existing layers in OpenStreetMap~\cite{OpenStreetMap} to extract $14,577$ and $6,910$ polygons for \textit{Building} and \textit{Road} layers, respectively. This increased our total annotated polygons from $11,382$ to $29,360$. 

To achieve better delineation of LULC classes, we generated hard \textit{Negative} examples as buffers of Building (with $3$ m buffer) and Road (with $5$ m buffer) classes, resulting a total of $51,618$ polygons as label examples (see Table~\ref{table:combined_lulc}). These labels are sparse and covered just $5.69\%$ of the pixels in the Maxar imagery, and they are used to train and test the models. The domain experts further annotated $1,219$ and $1,367$ polygons for the two main LULC classes of interest: Building and Crop labels, respectively - used as \textit{External} validation set. We excluded the \textit{Flooded Vegetation} class from our subsequent analysis due to the low quality of its corresponding label examples.

\begin{figure}[tbp]
    \centering
    \includegraphics[width=1.0\linewidth]{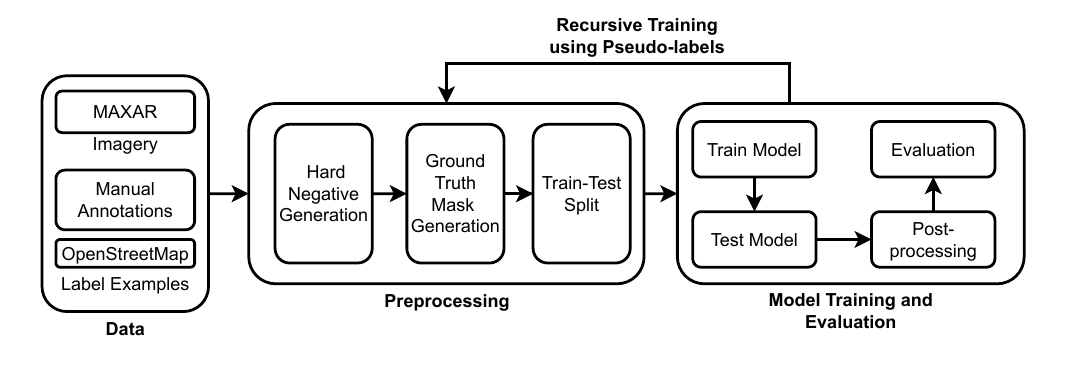}
    \caption{\textbf{Block diagram of our high-resolution teacher model.} Deep learning models are trained recursively using high-resolution Maxar imagery and label examples. We use non-overlapping train and test sets for training and testing the model, respectively.}\label{fig:teacher_model_overview}
\end{figure}

\section{Methodology}\label{sec:methodology}

Our methodology involves a knowledge transfer framework~\cite{wang2021knowledge, huang2022knowledge, beyer2022knowledge}, using \textit{High-Resolution Teacher Model} and \textit{Low-Resolution Student Model}. We describe the details of these models below.

\subsection{High-Resolution Teacher Model}
Figure~\ref{fig:teacher_model_overview} shows the block diagram of the high-resolution teacher model. The input data include high-resolution Maxar images and label examples for a portion of our AOI. The model produces a high-resolution LULC map. 
 We describe below the remaining steps: \textit{Preprocessing} and \textit{Model Training and Evaluation}. 

\paragraph*{Preprocessing}
Includes \textit{Hard Negative Generation}, \textit{Ground Truth Mask Generation} and \textit{Train-Test Split}.
We added a buffer zone for each of the \textit{Building} and \textit{Road} polygons to generate hard negative examples (see Sec.~\ref{subsec:annotations}). The added \textit{Negative} class minimizes the over-dominance of these two classes due to their over-representation in the training data and improves delineation of LULC classes. 
We generated a ground truth mask from the annotations by mapping each labeled pixel to its class index and assigning zero to the remaining unlabeled pixels. 
We split the input data into train and test sets by considering the high diversity across the vertical slices of the input Maxar imagery (see Fig.~\ref{fig:overview}), partly due to their different acquisition dates. For example, the left part of the imagery includes multiple cloud instances compared to the right part. As a result, we adopted a $70\%$ - $30\%$ vertical split of the data to train and test the model, respectively.

\paragraph*{Model Training and Evaluation} 
We adopted a deep learning-based semantic segmentation framework to implement the teacher model, comprising a U-Net~\cite{ronneberger2015u} with a ResNet-50~\cite{he2016deep} backbone pre-trained on the ImageNet~\cite{deng2009imagenet} dataset. We only considered errors from labeled pixels during training by ignoring errors from pixels with zero values in the mask.  We adopted a recursive training for our models, using the predicted labels from the initial model as pseudo-labels for the second round of training. The evaluation step involves testing the model's performance in detecting the LULC classes at a pixel level. We evaluated our approach using different validation sets: \textit{Whole} (using all the available label examples as shown in Table~\ref{table:combined_lulc}), \textit{Test} (using only the $30\%$ test split), and \textit{External} (using an external validation set collected for the top priority labels for the partner organization, i.e., Building and Crop).

\subsection{Low-Resolution Student Model}

The student model uses the publicly available Sentinel-2 images that cover the entire AOI. We used a set of label examples consisting of manually annotated labels, pseudo-labels from the teacher model, and labels from existing OpenStreetMap layers.
This results in $1,892,036$ polygons across eight LULC classes (see Table~\ref{table:combined_lulc}).  
During the generation of the ground truth mask for the student model, we down-sampled the high-resolution labels because the model works on the lower-resolution Sentinel-2 imagery.

We also implement the student model with deep learning-based semantic segmentation architectures, comprising U-Net~\cite{ronneberger2015u} with a backbone of a 5-layer fully connected convolutional neural network (FCN) provided in TorchGeo library~\cite{Stewart_TorchGeo_Deep_Learning_2022}. The student model setup uses the same $70\%$ - $30\%$ train-test split as the teacher model. 
The evaluation of the student model includes a post-processing step where the \textit{Road} and \textit{Building} LULC classes are merged back into the \textit{Built-up} class to be consistent with existing global maps for the comparison.

\section{Experimental Setup}\label{sec:experimental_setup}

In this Section, we present \textit{Baselines} (the existing global LULC maps used for comparison with our map), \textit{Training Setup} and \textit{Evaluation Setup and Metrics}.

\subsection{Baselines}\label{subsec:baselines}
We used existing global models as baselines to compare their corresponding LULC maps with our map produced with the student model. These existing maps are  Google's Dynamic World (GDW)~\cite{brown2022dynamic}, European Space Agency's (ESA) WorldCover~\cite{zanaga2022esa}, and Environmental Systems Research Institute's (ESRI) LULC~\cite{karra2021global}.
    GDW\cite{brown2022dynamic} is a near real-time land cover mapping tool that uses deep learning models and Sentinel-2 imagery. GDW provides global maps with $10$ LULC classes, updating every $2$ to $5$ days, aggregated using top-1 mode composite for the AOI in 2022.  
    ESA~\cite{zanaga2022esa}is a global land cover dataset derived from Sentinel-1 and Sentinel-2 data. ESA offers $10$ $\mathtt{m/pixel}$ resolution and produces detailed coverage of land types annually.  
    ESRI~\cite{karra2021global} is created using multi-source satellite imagery and contains a $10$ $\mathtt{m/pixel}$ resolution annually 2017-2023.

We used GDW's LULC types as the classes in our work except for the \textit{Flooded Vegetation} class. These were excluded due to quality issues observed in the manually annotated polygons by domain experts. We aligned classes from different maps based on their definitions. To maintain a consistent set of LULC classes across the maps, we excluded \textit{Snow and Ice}, \textit{Herbaceous Wetland}, \textit{Mangroves}, and \textit{Moss and Lichen} labels from ESA's map, and \textit{Snow/Ice}, \textit{Clouds}, \textit{Herbaceous Wetland}, \textit{Mangroves}, \textit{Moss and Lichen}, and \textit{Shadow} from ESRI's map. During evaluation, the excluded classes were relabeled as \textit{Others}.

\subsection{Training Setup}\label{subsec:training_setup}
We adopted a similar training setup for our teacher and student models, involving \textit{class weighting}, patch size = $512$, batch size = $32$, minimum epochs = $100$, maximum epochs = $300$, learning rate = $0.0003$, and the cross-entropy loss. 
We used a sequence of augmentation steps including $90\deg$ and $225\deg$ rotations, horizontal and vertical flips - all with a probability of $p=0.5$.

\subsection{Evaluation Setup and Metrics}\label{subsec:evaluation_metrics}
 During evaluation, pixels predicted as the Negative class with the highest probability are relabeled with the second-highest probable class. To ensure a consistent set of LULC classes across maps, we merged the \textit{Building} and \textit{Road} classes into a single \textit{Built-up} class during the evaluation of the student model output.
The metrics include Accuracy, Precision, Recall, $F_1$ score and Intersection over Union (IoU). We computed these metrics for each LULC class, using one-vs-all strategy, and reported the Macro average. We used confusion matrices to understand the misclassification of pixels across the LULC classes. We also used an agreement matrix as a measure of consistency across the LULC maps.

\section{Results}\label{sec:results}

\begin{figure}[htbp] 
    \centering
\includegraphics[width=1.0\linewidth]{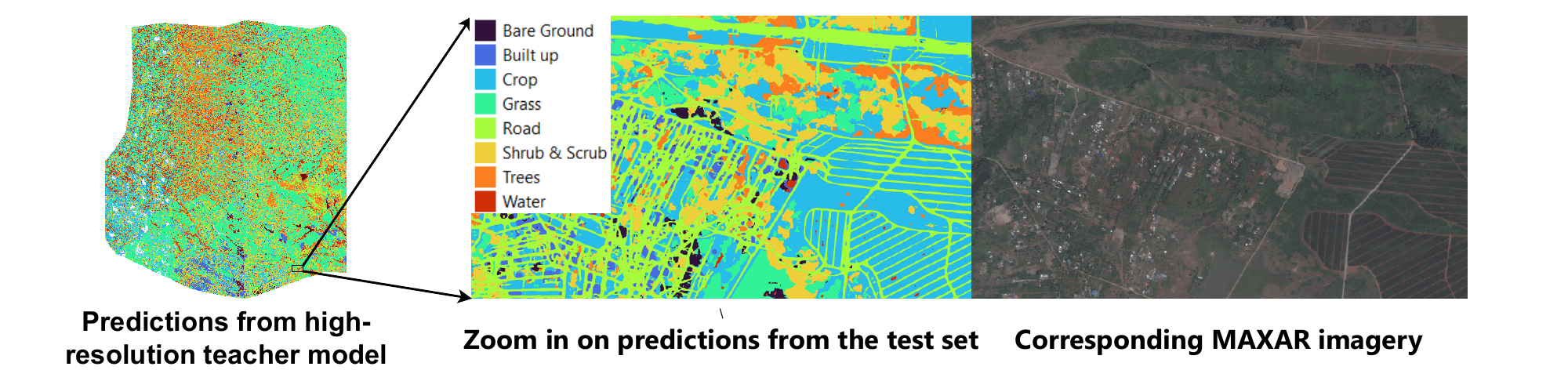}
    \caption{\textbf{High-resolution LULC map generated using the teacher model.} Zoomed-in version of the LULC map from the test set, which was not seen during training, shows a high-quality map with clear delineation.}\label{fig:maxar_maps}
\end{figure}
\subsection{High-Resolution Teacher Model}\label{subsec:high_res_maps}
The teacher model produces a high-resolution LULC map for a portion of our AOI (see Fig.~\ref{fig:maxar_maps}). A subsequent zoom on imagery from the held-out test set shows a well-delineated classification of LULC classes. This result supports our approach to treat \textit{Building} and \textit{Road} LULC classes as separate classes by reducing ambiguity and improving delineation. 
%
%
\begin{figure}[htbp]  
    \centering

      \begin{subfigure}[b]{0.45\linewidth}
        \centering
        \includegraphics[width=\linewidth]{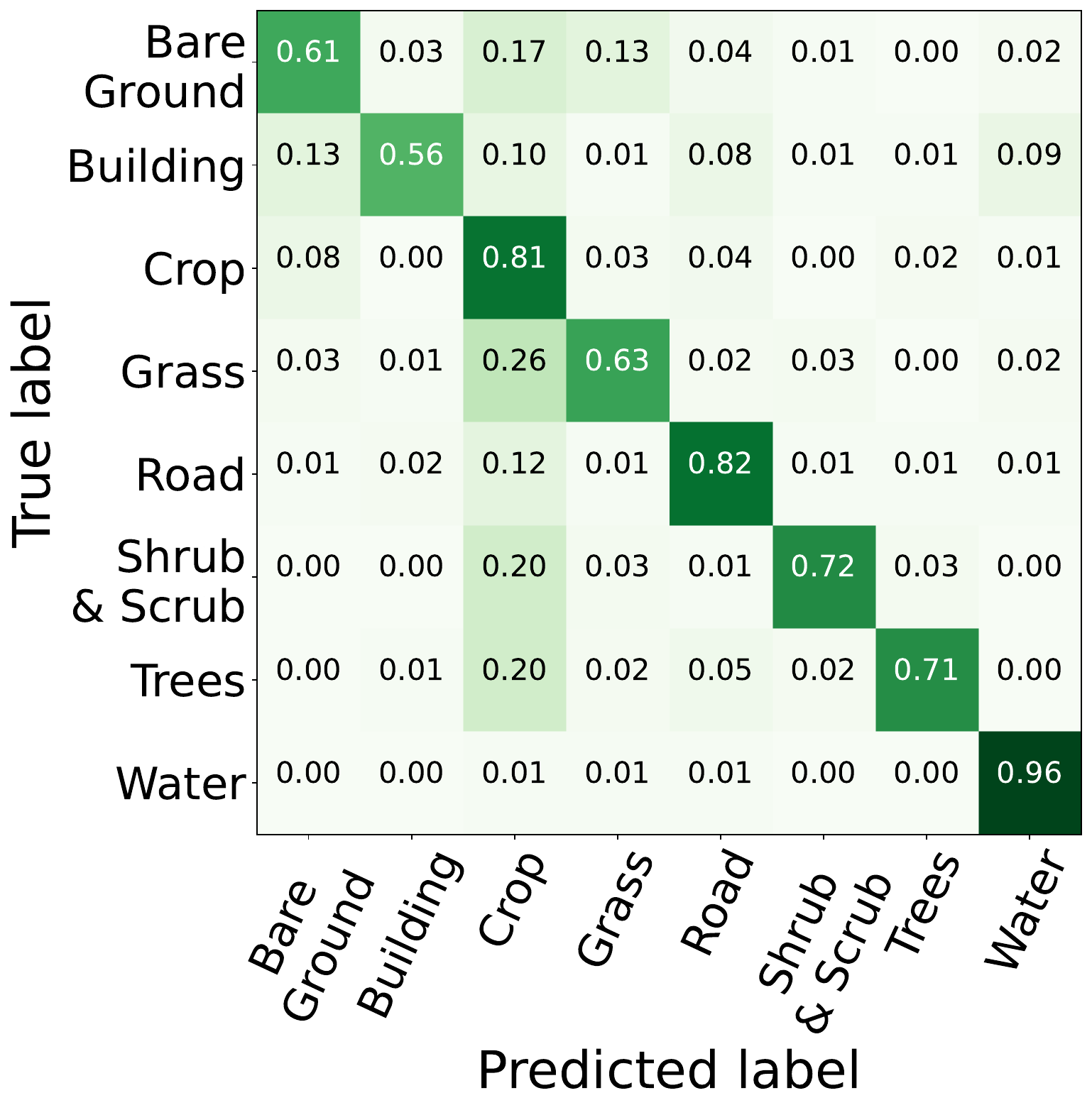}
        \caption{Whole set}
    \end{subfigure}
    \begin{subfigure}[b]{0.45\linewidth}
        \centering
        \includegraphics[width=\linewidth]{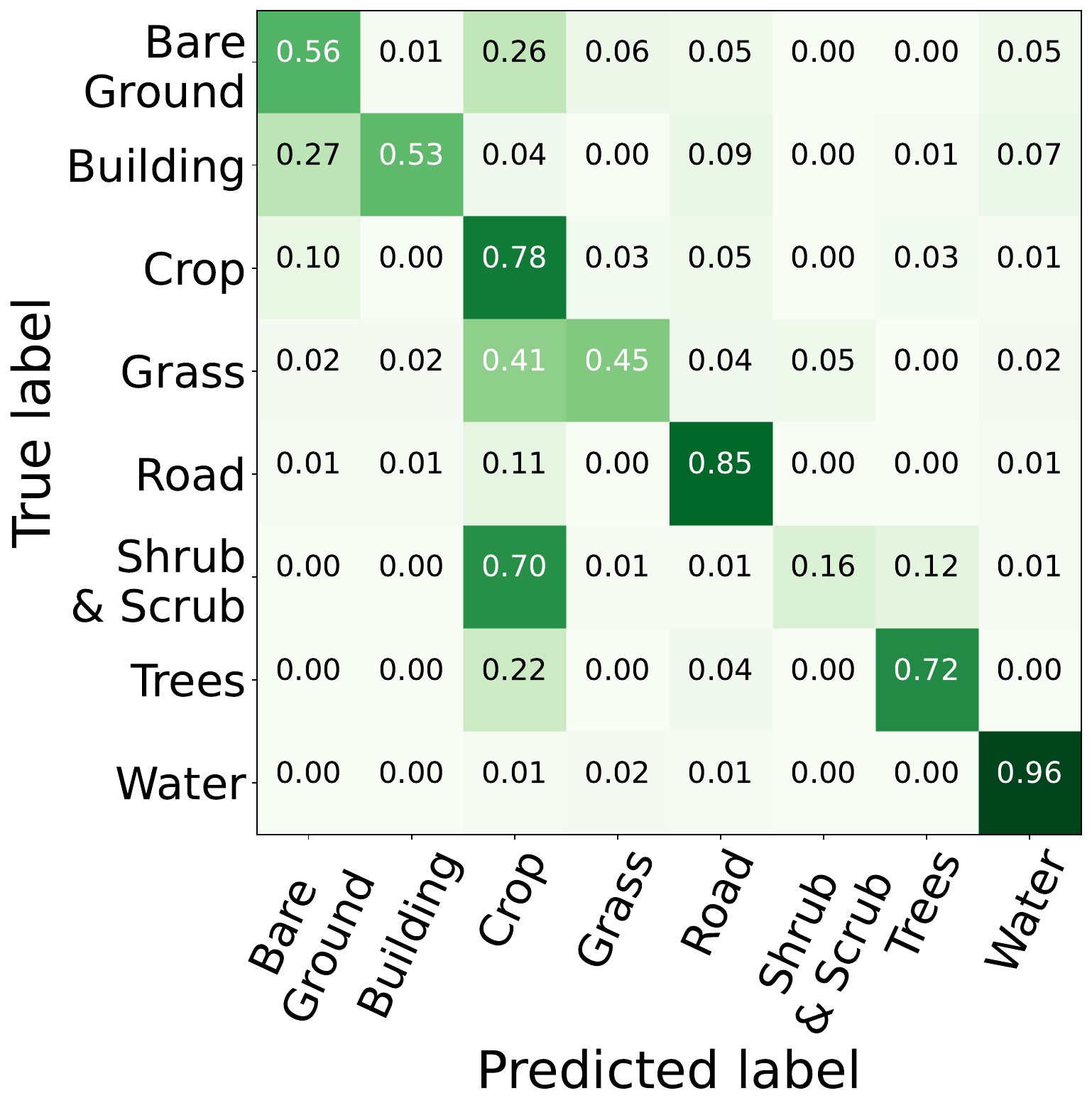}
        \caption{Test set}
    \end{subfigure}
    \begin{subfigure}[b]{0.45\linewidth}
        \centering
        \includegraphics[width=\linewidth]{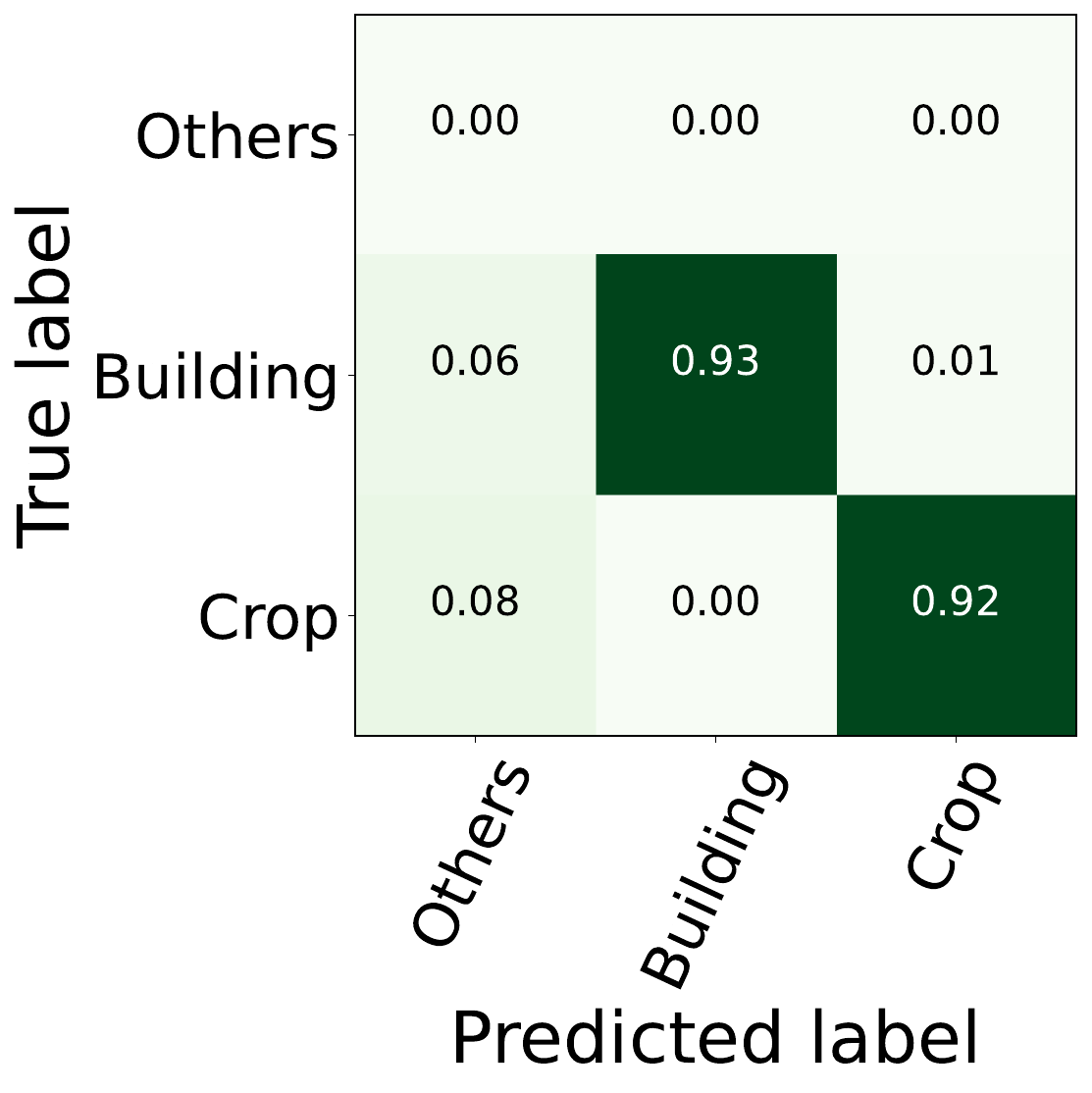}
        \caption{External set}
    \end{subfigure}     
    \caption{\textbf{Confusion matrices of LULC maps from the high-resolution teacher model across (a) Whole, (b) Test and (c) External sets.}}\label{fig:maxar_conf_matrices}  
\end{figure}  

The confusion matrices in Fig.~\ref{fig:maxar_conf_matrices} show the classification accuracy among the classes, while also highlighting the misclassifications of \textit{Bare Ground}, \textit{Grass}, \textit{Trees}, and \textit{Shrubs} as \textit{Crop}.
The results derived from the Test set follow the trend observed in the Whole set, but the confusion between \textit{Shrub \& Scrub} and \textit{Crop} classes worsens. The performance of detecting the priority classes: \textit{Building} and \textit{Road}, using the External set, is even higher as shown in Fig.~\ref{fig:maxar_conf_matrices} (c). Table~\ref{table:maxar_aggregated} shows the metrics of the teacher model when evaluated on different validation sets. 
The results are encouraging for the Test set, while the performance on the External set is even higher with $0.96\pm0.0$ and $0.86\pm0.01$ of $F_1$ score and IoU, respectively.

\begin{table*}[tbp]  
\centering  
\caption{\textbf{Performance evaluation of our high-resolution teacher model}. Evaluation was done using different validation sets: Whole, Test, and External. IoU: Intersection over Union. The standard deviation of the metrics indicates variations across the classes.
}\label{table:maxar_aggregated}
\resizebox{0.6\linewidth}{!}{
\begin{tabular}{llllll}
\toprule
Validation Set & Accuracy & Precision & Recall & $F_1$ score & IoU \\
\midrule
Whole & 0.95±0.06 & 0.44±0.32 & 0.73±0.13 & 0.50±0.27 & 0.59±0.18 \\
Test & 0.94±0.06 & 0.40±0.42 & 0.62±0.26 & 0.40±0.36 & 0.5±0.27 \\
External & 0.96±0.02 & 1.0±0.0 & 0.92±0.01 & 0.96±0.0 & 0.86±0.01 \\
\bottomrule
\end{tabular}
}
\end{table*}
\begin{figure}[tbp]  
    \centering  
    \begin{subfigure}[b]{0.49\linewidth}
        \centering
        \includegraphics[width=\linewidth]{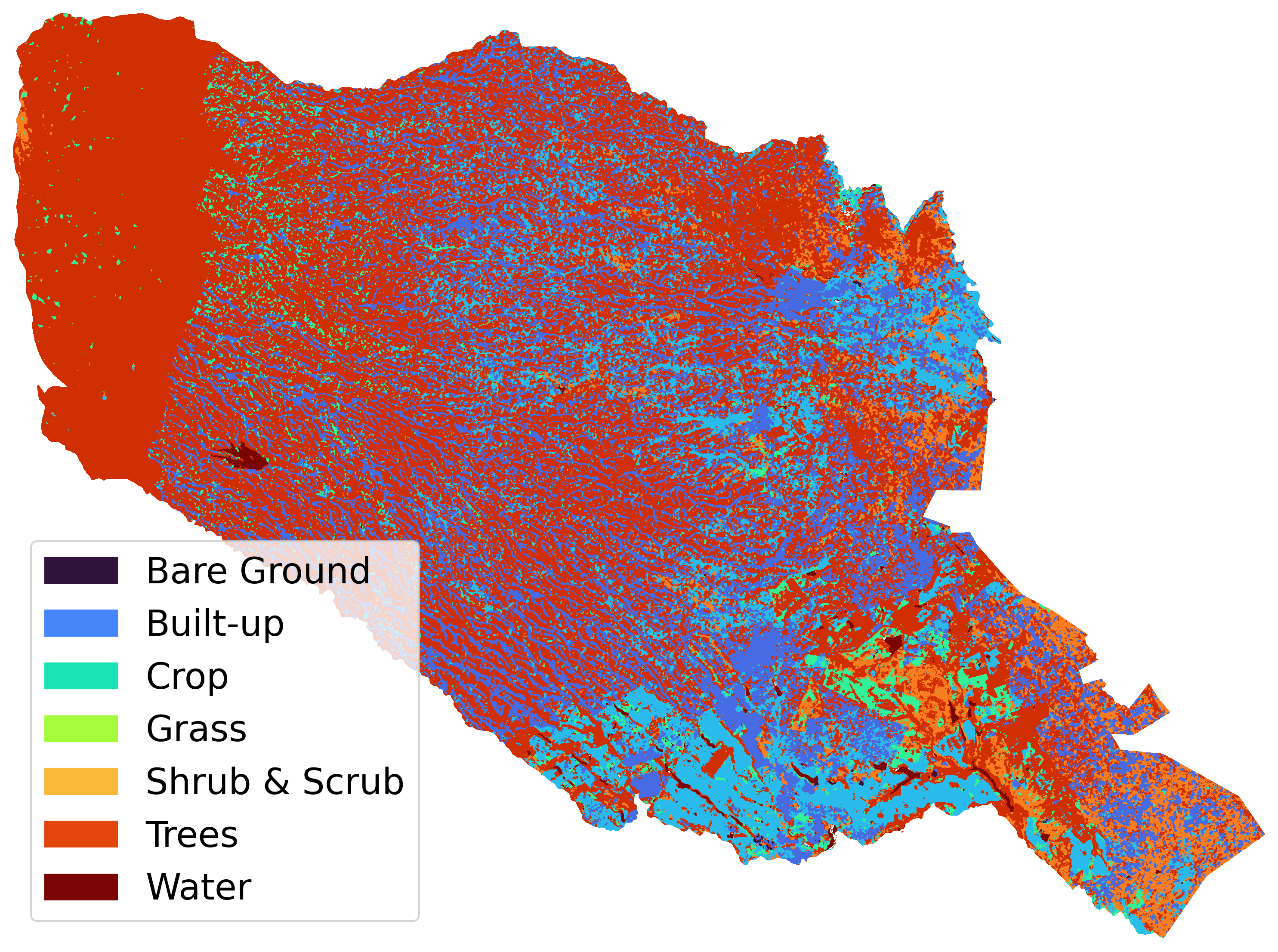}
        \caption{GDW~\cite{brown2022dynamic}}
    \end{subfigure}
    \begin{subfigure}[b]{0.49\linewidth}
        \centering
        \includegraphics[width=\linewidth]{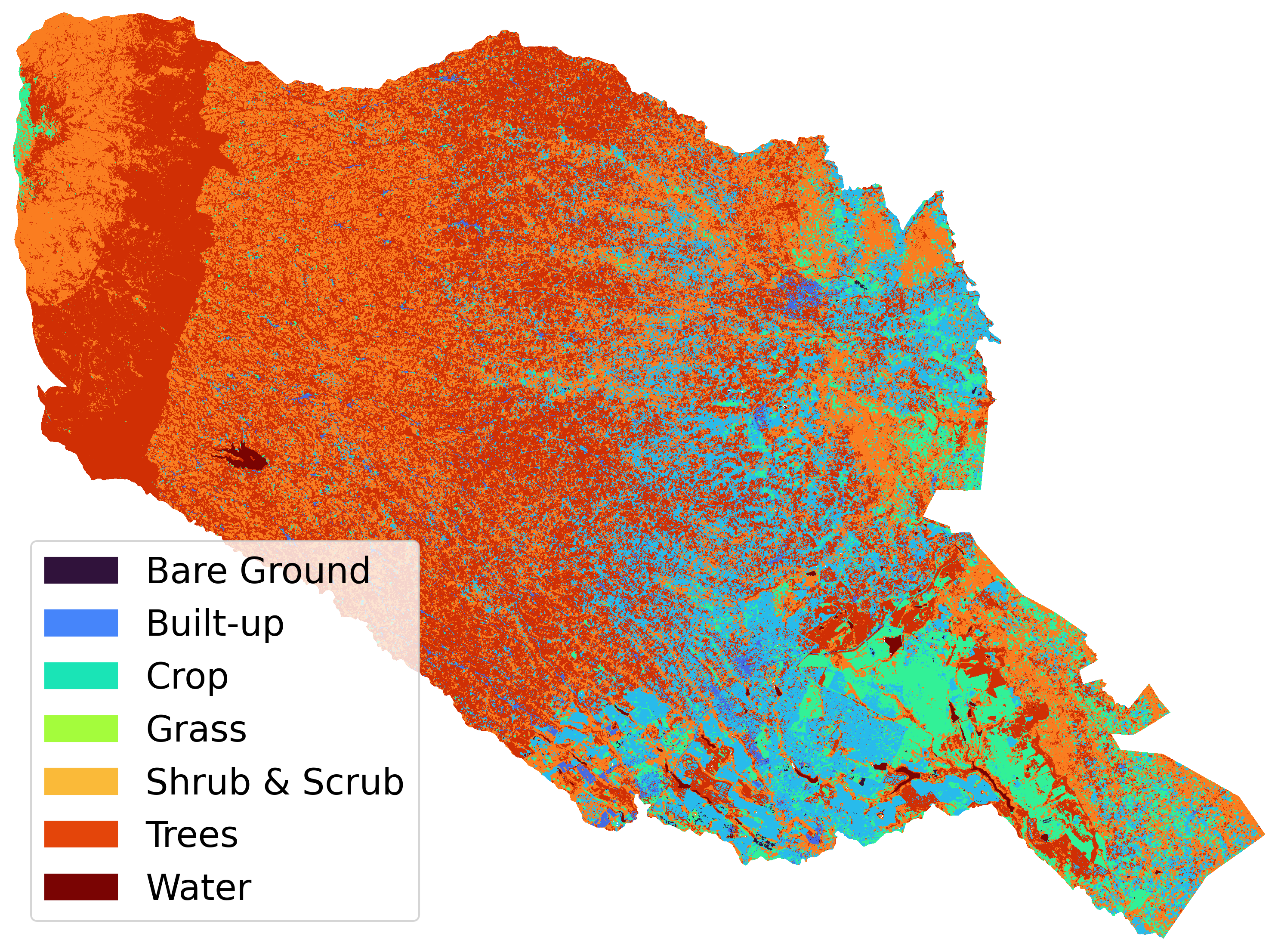}
        \caption{ESA~\cite{zanaga2022esa}}
    \end{subfigure}
    
    \begin{subfigure}[b]{0.49\linewidth}
        \centering
        \includegraphics[width=\linewidth]{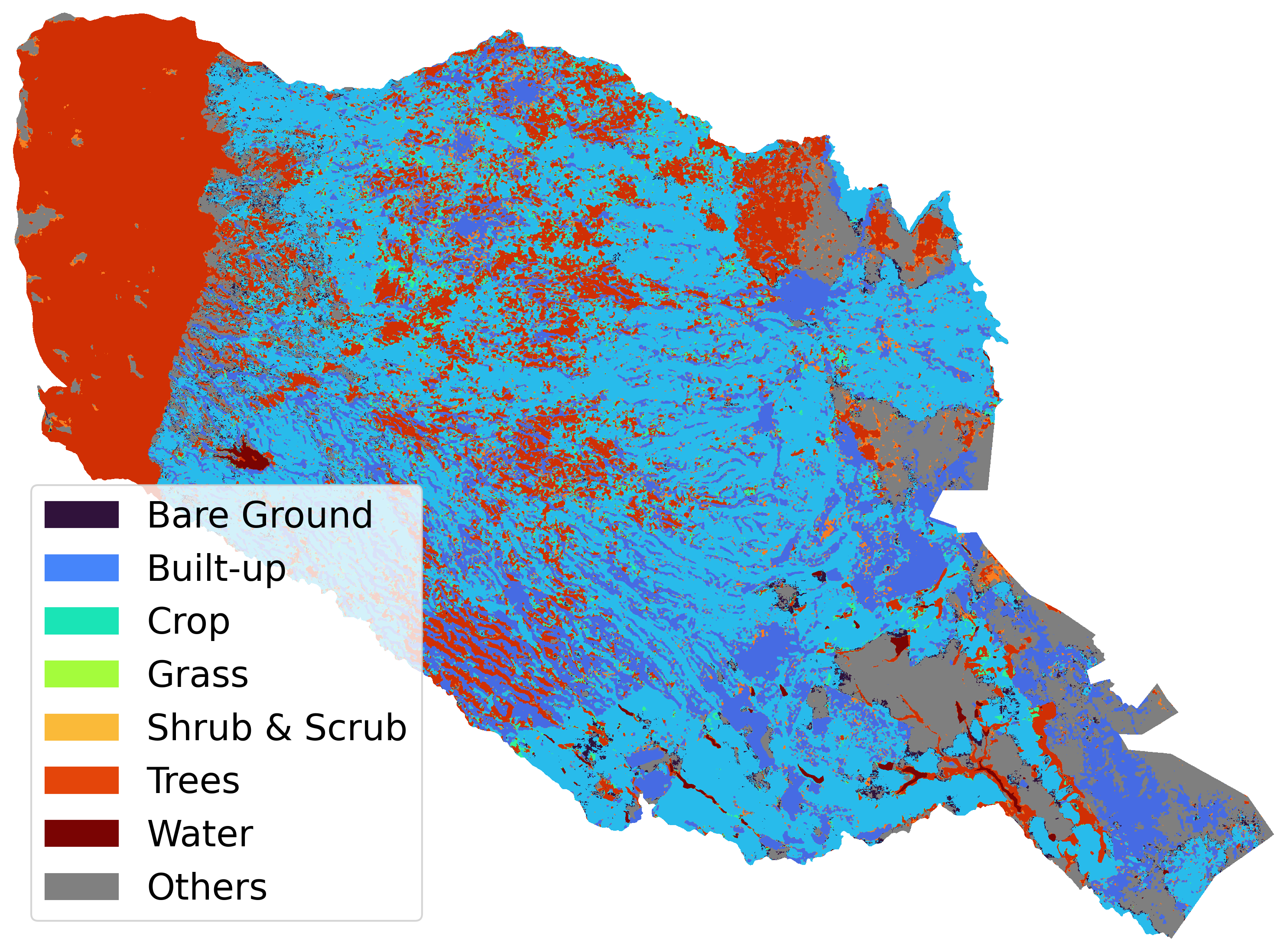}
        \caption{ESRI~\cite{karra2021global}}
    \end{subfigure}
    \begin{subfigure}[b]{0.49\linewidth}
        \centering
        \includegraphics[width=\linewidth]{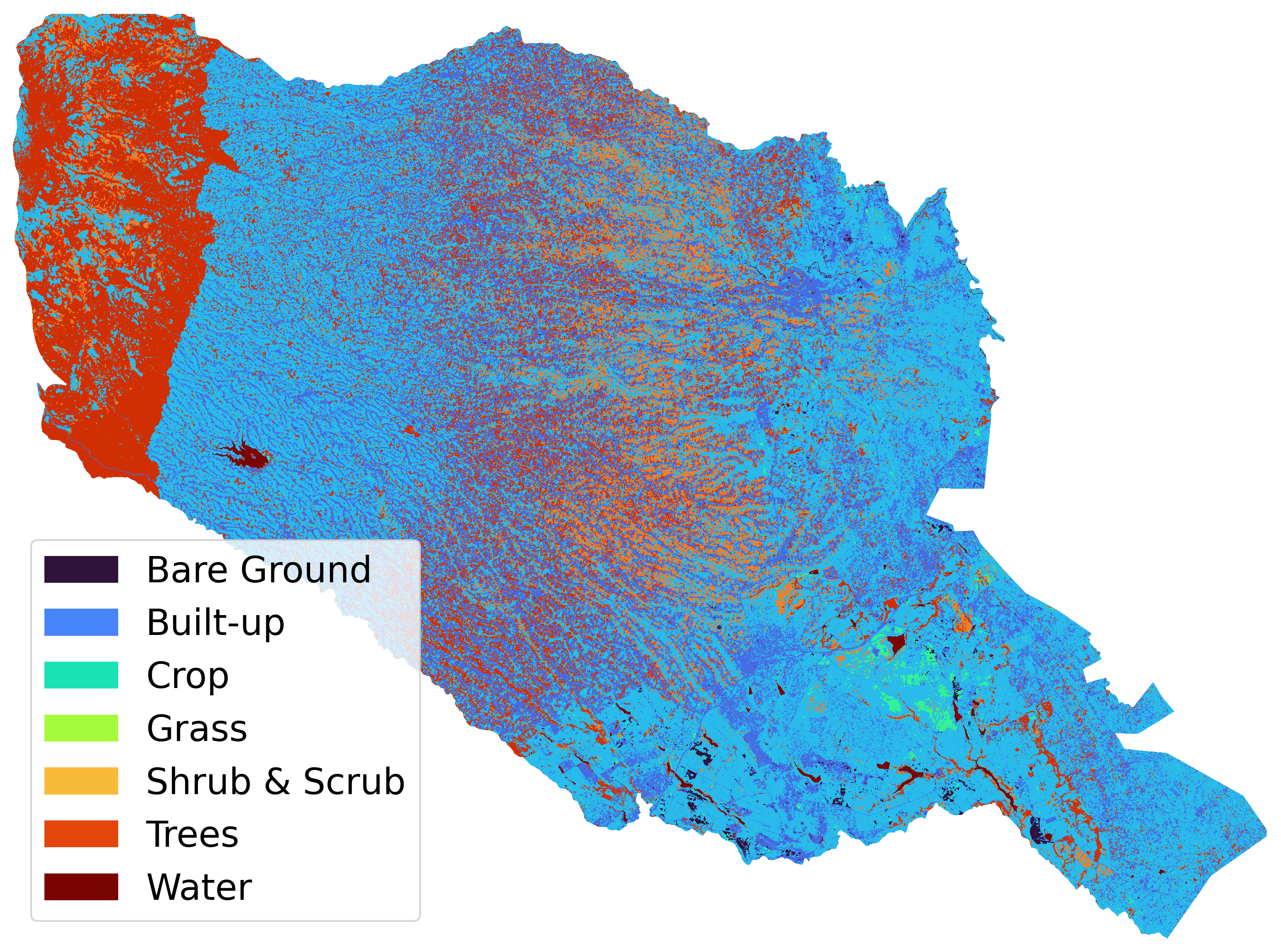}
        \caption{\ourmethod (ours)}
    \end{subfigure}
      
    \caption{\textbf{Comparison of LULC maps across Murang'a county}. (a) GDW~\cite{brown2022dynamic}, (b) ESA~\cite{zanaga2022esa}, (c) ESRI~\cite{karra2021global}, and (d) \ourmethod. Both the ESRI and \ourmethod maps demonstrate similar patterns, such as a higher observation of croplands. Overall, the \ourmethod map exhibits higher quality compared to the global maps (a) - (c).}  
      
    \label{fig:merged_maps}  
\end{figure}  

    

\begin{figure}[htbp!]
\centering
    \begin{subfigure}[b]{0.35\linewidth}
        \centering
        \includegraphics[width=\linewidth]{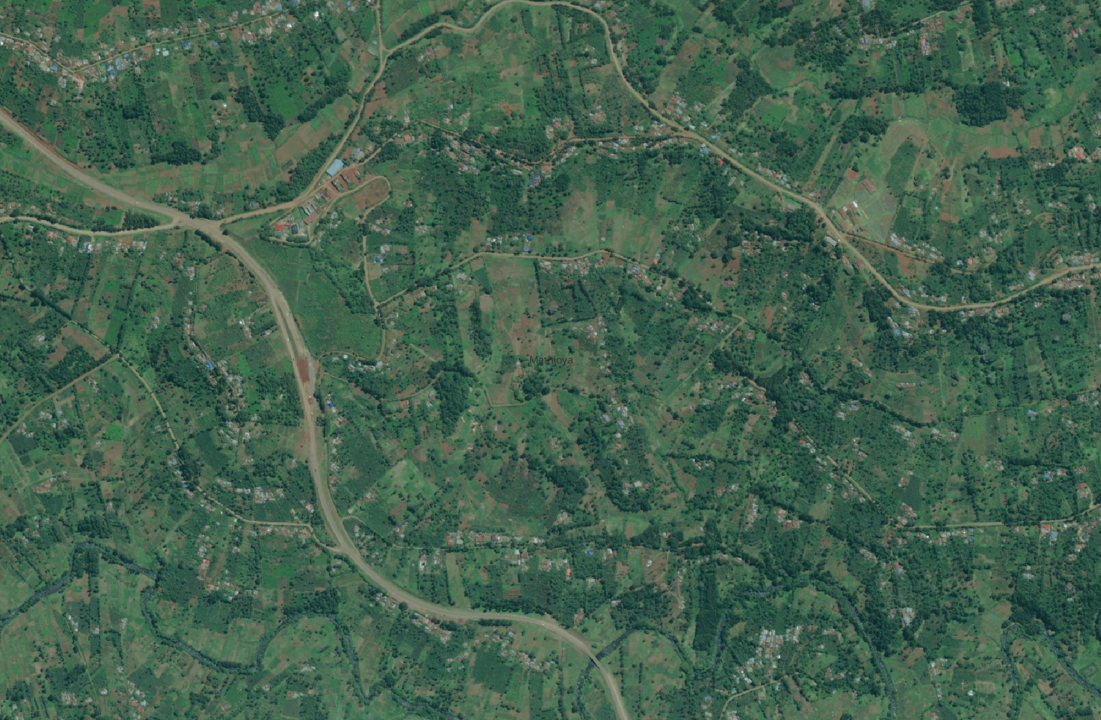}
        \caption{Bing Map - Mathioya, Murang'a  county}
    \end{subfigure}
        \begin{subfigure}[b]{0.35\linewidth}
        \centering
        \includegraphics[width=\linewidth]{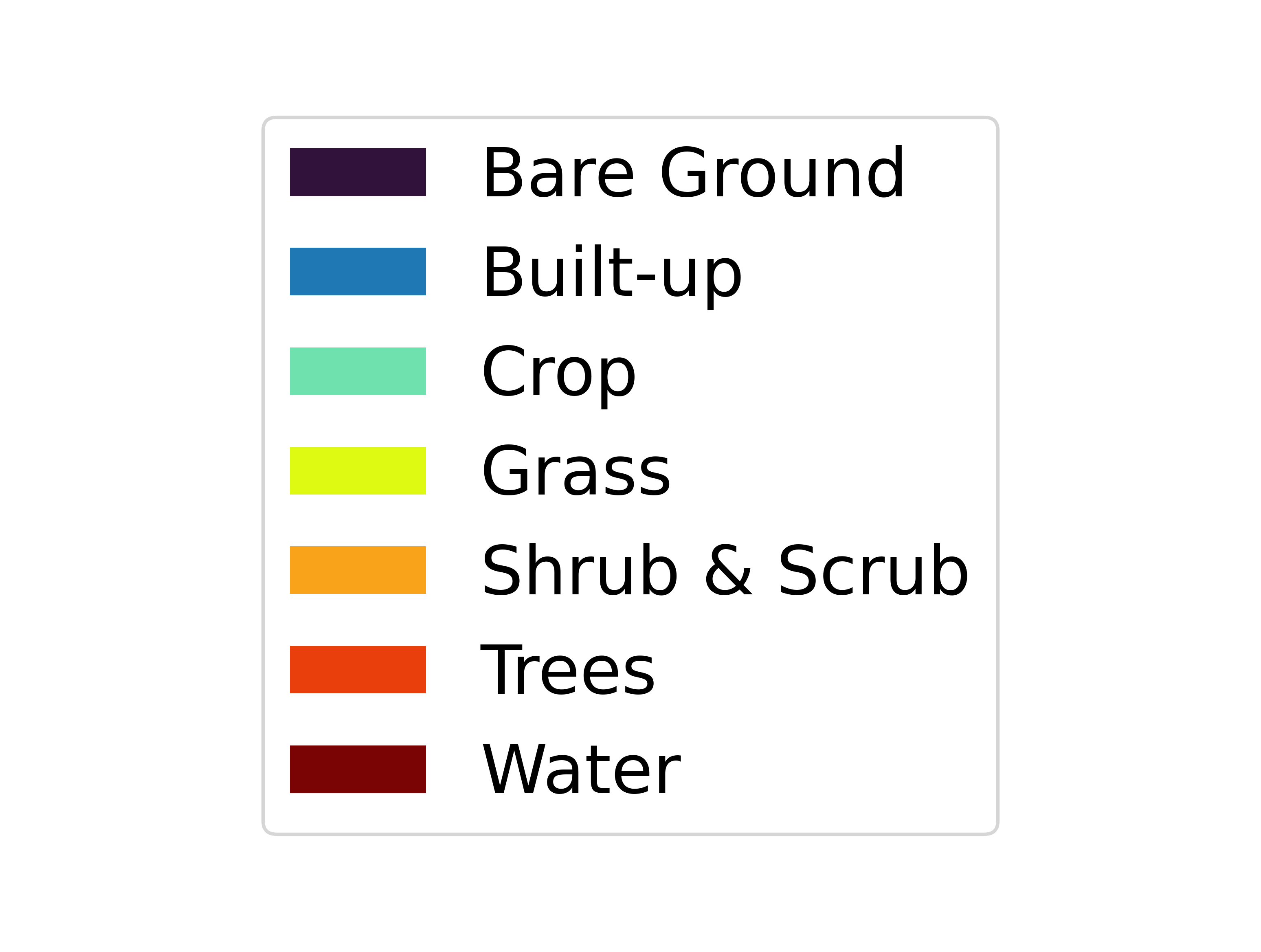}
        \caption{Legend of LULC classes}
    \end{subfigure}
    \begin{subfigure}[b]{0.35\linewidth}
        \centering
        \includegraphics[width=\linewidth]{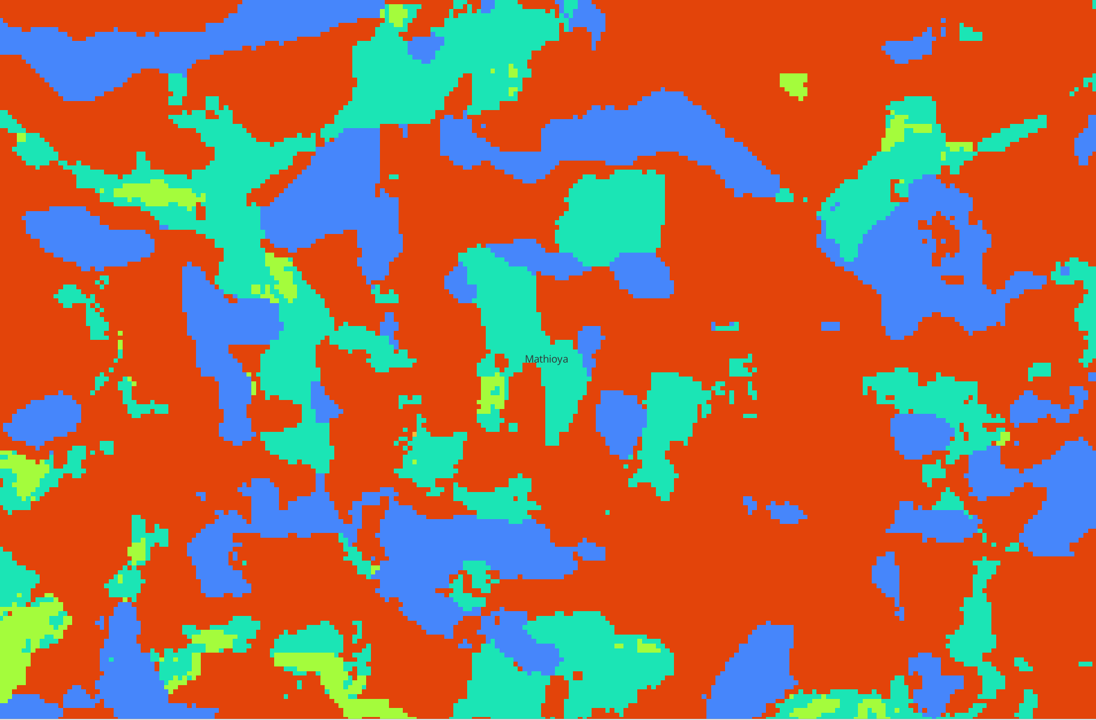}
        \caption{GDW~\cite{brown2022dynamic}}
    \end{subfigure}
    \begin{subfigure}[b]{0.35\linewidth}
        \centering
        \includegraphics[width=\linewidth]{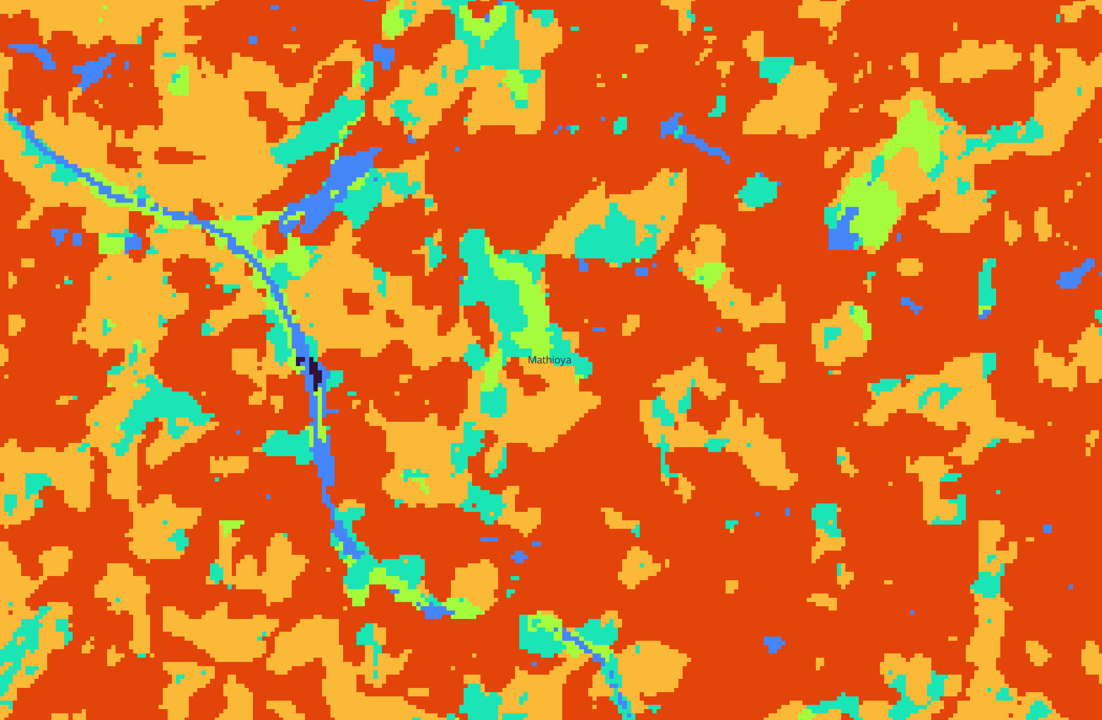}
        \caption{ESA~\cite{zanaga2022esa}}
    \end{subfigure}
    \begin{subfigure}[b]{0.35\linewidth}
        \centering
        \includegraphics[width=\linewidth]{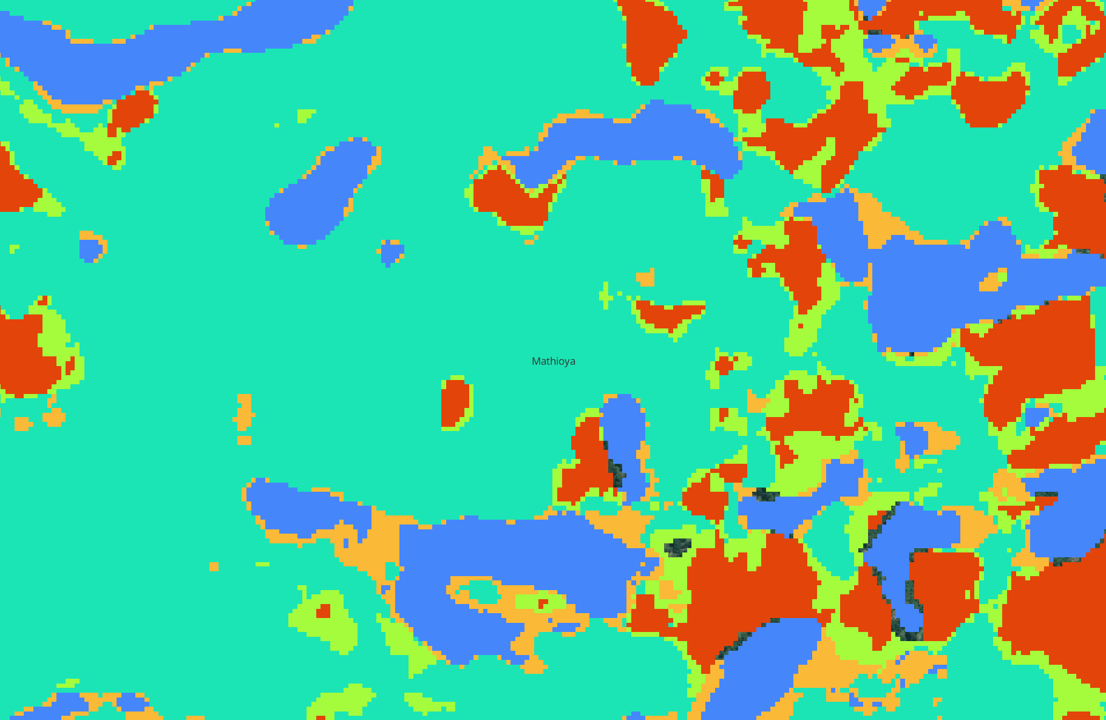}
        \caption{ESRI~\cite{karra2021global}}
    \end{subfigure}
    \begin{subfigure}[b]{0.35\linewidth}
        \centering
        \includegraphics[width=\linewidth]{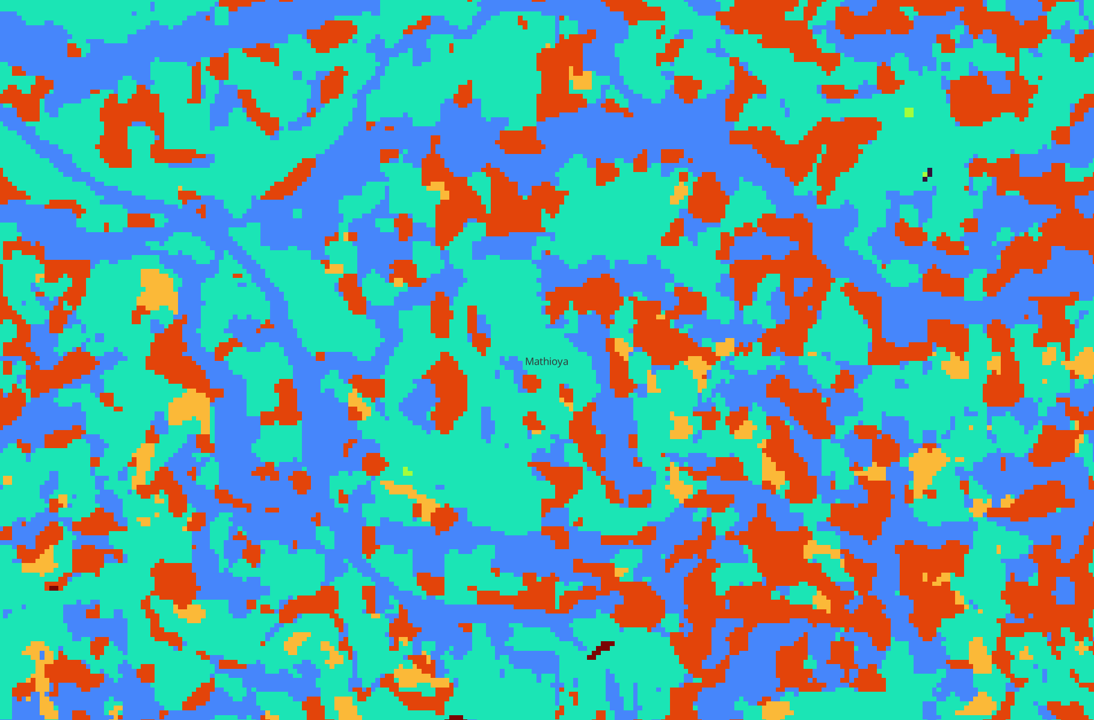}
        \caption{\ourmethod (ours)}
    \end{subfigure}
\caption{\textbf{Zoomed-in comparison of the LULC maps around Mathioya town in Murang'a county}: (a) Bing imagery, (b) legend of LULC classes, (c) GDW\cite{brown2022dynamic}, (d) ESA\cite{zanaga2022esa}, (e) ESRI~\cite{karra2021global}, and (f) \ourmethod from our local model exhibits higher quality compared to the global maps (c) - (e).}
\label{fig:map-examples-mathioya}
\end{figure}

\subsection{Low-Resolution Student Model}\label{subsec:low_res_maps}
The advantage of our student model is its ability to generate the LULC map for the entire AOI, i.e., the Murang'a county of Kenya, as it uses publicly available Sentinel-2 images.
Figure~\ref{fig:merged_maps} shows different LULC maps of the AOI including 
the map from the student model of the proposed \ourmethod and Baseline maps: GDW~\cite{brown2022dynamic}, ESA~\cite{zanaga2022esa} and ESRI~\cite{karra2021global}. The GDW and ESA maps, shown in Fig.~\ref{fig:merged_maps} (a) and (b), respectively, have similar trends of overestimation of \textit{Trees}, particularly in the northern and eastern parts of the county. The ESA map further shows an overestimation of the \textit{Shrub \& Scrub} class across the county, while the \textit{Built-up} class is underestimated. On the other hand, the ESRI and our (\ourmethod) maps, shown in Fig.~\ref{fig:merged_maps} (c) and (d), respectively, exhibit similar trends, including widespread \textit{Crop} and \textit{Built-up} classes. The \ourmethod map demonstrates high-quality patterns, including the \textit{Built-up} class. The \textit{Others} class in ESRI's map includes labels that were dropped from the evaluation, such as Flooded Vegetation, Snow/Ice, Clouds and Shadow, as described in Sec.~\ref{sec:experimental_setup}. 
Figure~\ref{fig:map-examples-mathioya} shows the closer look of these maps zoomed in around Mathioya town, in northern part of the county, known for its agricultural activities. The qualitative comparison of the maps further confirms the patterns observed in Fig.~\ref{fig:merged_maps}. The existing GDW, ESA and ESRI maps, produced by the corresponding global models, shown in Fig.~\ref{fig:map-examples-mathioya} (c) - (e), exhibit lower quality compared to the \ourmethod map, produced using our local model, shown in Fig.~\ref{fig:map-examples-mathioya} (f). The GDW and ESA maps predominantly show \textit{Trees}, whereas the ESA map distinctively shows a higher observation of \textit{Shrub \& Scrub} around Mathioya. The ESRI map, shown in Fig.~\ref{fig:map-examples-mathioya} (e), indicates an overestimation of Croplands while Trees and Built-up areas are underestimated. 
Our \ourmethod map, shown in Fig.~\ref{fig:map-examples-mathioya} (f), presents a balanced view of \textit{Crop}, \textit{Trees} and \textit{Built-up} classes. All the existing global maps show poor quality in mapping \textit{Built-up} instances such as buildings and roads. Even the highway from northeast to south Mathioya, see Fig.~\ref{fig:map-examples-mathioya} (a), was rarely detected by the global models.
Existing LULC maps also fail to correctly map buildings that follow roads across the region.


Table~\ref{table:sentinel_aggregated} shows the evaluation metrics computed from the Whole set and averaged across the seven LULC classes considered. The results demonstrate that our map achieves the highest performance across all the evaluation metrics, with improvements of $0.14$ in the $F_1$ score and $0.21$ in Intersection-over-Union, compared to the map produced by the best global model, i.e., ESRI~\cite{karra2021global}.
Table~\ref{table:sentinel_external_results} shows the maps evaluated on the External set. The results once again confirm the superior quality of our map over the existing global maps.

\begin{table*}[tbp!]
    \centering
    \caption{\textbf{Comparison of maps on the Whole validation set}. The maps include \ourmethod map - derived from our local student model and existing maps: GDW~\cite{brown2022dynamic}, ESA~\cite{zanaga2022esa} and ESRI~\cite{karra2021global}, produced by global models. The values represent pixel-level performance metrics aggregated over all LULC classes using Macro average and standard deviation ($mean\pm std $). The standard deviation of the metrics indicates variations across the classes.
}\label{table:sentinel_aggregated}
    \resizebox{0.6\linewidth}{!}{
\begin{tabular}{llllll}
\toprule
\textbf{LULC Map} &\textbf{ Accuracy} &\textbf{ Precision} & \textbf{Recall} &\textbf{ $F_1$ score} & \textbf{IoU} \\
\midrule
GDW~\cite{brown2022dynamic} & 0.83±0.2 & 0.25±0.37 & 0.26±0.24 & 0.19±0.25 & 0.17±0.17 \\
ESA~\cite{zanaga2022esa} & 0.76±0.23 & 0.16±0.27 & 0.18±0.2 & 0.08±0.17 & 0.11±0.12 \\
ESRI~\cite{karra2021global} & 0.88±0.15 & 0.33±0.42 & 0.32±0.34 & 0.29±0.34 & 0.25±0.3 \\
\ourmethod (ours) & \textbf{0.92±0.1} & \textbf{0.42±0.42} & \textbf{0.58±0.3} & \textbf{0.43±0.37} & \textbf{0.46±0.3 }\\
\bottomrule
\end{tabular}
}
\end{table*}

\begin{table*}[tbp!]  
\centering  
\caption{\textbf{Comparison of maps on the External validation set consisting of \textit{Building} and \textit{Crop} classes}. The maps include \ourmethod map - derived from our local student model and existing maps: GDW~\cite{brown2022dynamic}, ESA~\cite{zanaga2022esa} and ESRI~\cite{karra2021global}, produced by global models. Our local model achieved the highest value across the all the metrics and classes considered.} \label{table:sentinel_external_results}  
\resizebox{0.7\linewidth}{!}{
\begin{tabular}{llccccc}  
\toprule 
\textbf{LULC Map} & \textbf{LULC class} & \textbf{Accuracy} & \textbf{Precision} & \textbf{Recall} & \textbf{$F_1$ score} & \textbf{IoU} \\  \midrule
\multirow{2}{*}{GDW~\cite{brown2022dynamic}} & Built-up & 0.472 & 0.403 & 0.737 & 0.521 & 0.584 \\    
 & Crop & 0.416 & 0.664 & 0.09 & 0.158 & 0.047 \\  \midrule
\multirow{2}{*}{ESA~\cite{zanaga2022esa}} & Built-up & 0.612 & 0.516 & 0.029 & 0.055 & 0.015 \\   
 & Crop & 0.46 & 0.566 & 0.491 & 0.526 & 0.325 \\  \midrule
\multirow{2}{*}{ESRI~\cite{karra2021global}} & Built-up & 0.665 & 0.543 & 0.892 & 0.675 & 0.806 \\    
 & Crop & 0.652 & 0.904 & 0.481 & 0.628 & 0.317 \\ \midrule
\multirow{2}{*}{\ourmethod (ours)} & Built-up & \textbf{0.898} & \textbf{0.82} & \textbf{0.945} & \textbf{0.878} & \textbf{0.896} \\ 
 & Crop & \textbf{0.891} & \textbf{0.963} & \textbf{0.853} & \textbf{0.905} & \textbf{0.744} \\ \bottomrule 
\end{tabular}  
}
\end{table*}

\section{Discussion}\label{sec:discussion}

\subsection{Inconsistencies and Reduced Accuracy Highlight Limitations of Global LULC Maps}\label{subsec:agreement} 
\begin{figure}[htbp]
    \centering
    \includegraphics[width=0.45\linewidth]{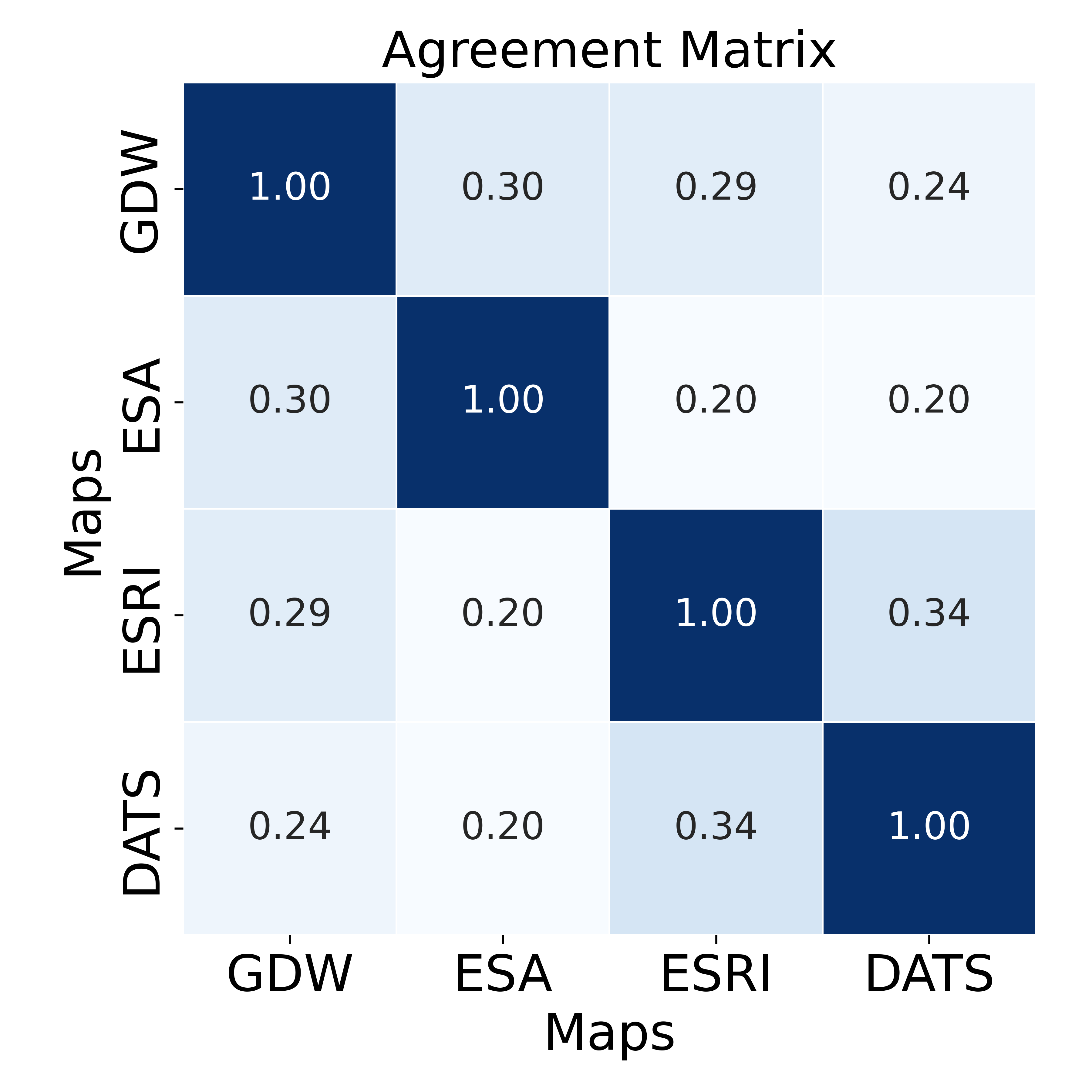}
    \caption{\textbf{Existing global maps exhibit a low level of agreement among themselves.} The result highlights both lower accuracy and inconsistencies of global maps in local African contexts. The map from \ourmethod (ours) achieves the highest agreement with ESRI.
    }\label{fig:aggreement_matrix}
\end{figure}

 Venter et al.~\cite{venter2022global} reported higher accuracy and agreement-level of existing global LULC maps when validated globally and in Europe. But when we validated these maps in the Murang'a county of Kenya in Africa, they exhibit poor performance (see Sec.~\ref{sec:results}) and inconsistencies (see Fig.~\ref{fig:aggreement_matrix}). Our map achieves the highest agreement rate of $0.34$ with the ESRI~\cite{karra2021global} map, followed by $0.24$ with the GDW~\cite{brown2022dynamic} map. The ESA~\cite{zanaga2022esa} map, which tends to over-predict the \textit{Trees} and \textit{Shrub \& Scrub} classes as shown in Fig.~\ref{fig:merged_maps} (b), has the least agreement with the remaining maps. This also further supports the limitations of these global maps reported in prior research~\cite{kerner2024accurate}.

Table~\ref{table:area_coverage} highlights the variability in area coverages for the LULC classes across the maps. For example, the \textit{Built-up} class covers a significant area except in the ESA~\cite{zanaga2022esa} map, with our map indicating a coverage of $529.64$ km\(^2\) (representing $20.96\%$ of the county), similar to a coverage of $501.71$ km\(^2\) ($19.86\%$) in the GDW~\cite{brown2022dynamic} map. The ESA~\cite{zanaga2022esa} map underestimates the \textit{Built-up} class, with only $42.55$ km\(^2\) ($1.68\%$) coverage. GDW underestimates \textit{Crop} coverage, whereas both the ESRI and \ourmethod maps show similar coverage distribution. 
 We noted discrepancies in the estimations of \textit{Trees} coverage across the maps.
 GDW map's $1461.64$ km\(^2\) ($58.85\%$), compared to our map's $510.92$ km\(^2\) ($20.22\%$), indicates an overestimation by GDW~\cite{brown2022dynamic} —a pattern consistently observed in Fig.~\ref{fig:merged_maps} and Fig.~\ref{fig:map-examples-mathioya}. The ESA~\cite{zanaga2022esa} map shows a $844.24$ km\(^2\) ($33.42\%$) coverage of \textit{Shrub \& Scrub}, which is an overestimation compared to the remaining maps.

\begin{table*}[htbp]
    \centering
    \caption{\textbf{Comparison of area coverages of LULC classes in Murang'a county.} The areas were estimated from maps produced by exiting global models and our local student model (\ourmethod). The global maps include GDW~\cite{brown2022dynamic}, ESA~\cite{zanaga2022esa} and ESRI~\cite{karra2021global}).}\label{table:area_coverage}
    \resizebox{0.8\linewidth}{!}{
\begin{tabular}{llrrrrrrrr}
\toprule
& \multicolumn{8}{c}{\textbf{Maps}} \\
   & \multicolumn{2}{c}{\textbf{GDW~\cite{brown2022dynamic}}} & \multicolumn{2}{c}{\textbf{ESA~\cite{zanaga2022esa}}} & \multicolumn{2}{c}{\textbf{ESRI~\cite{karra2021global}}} & \multicolumn{2}{c}{\textbf{\ourmethod} (ours)}\\
   \cmidrule(lr){2-3} \cmidrule(lr){4-5} \cmidrule(lr){6-7} \cmidrule(lr){8-9}
\textbf{LULC class} & $km^2$ & $\%$ & $km^2$ & $\%$ & $km^2$ & $\%$ & $km^2$ & $\%$ \\
\midrule
 Bare Ground & 1.03 & 0.04 & 2.74 & 0.11 & 39.20 & 1.55 & 10.01 & 0.40 \\
 Built-up & 501.71 & 19.86 & 42.55 & 1.68 & 443.21 & 17.54 & 529.64 & 20.96 \\
 Crop & 335.99 & 13.30 & 431.04 & 17.06 & 1019.88 & 40.37 & 1290.75 & 51.09 \\
 Grass & 80.91 & 3.20 & 202.61 & 8.02 & 77.78 & 3.08 & 14.32 & 0.57 \\
 Shrub \& Scrub & 131.23 & 5.19 & 844.24 & 33.42 & 74.39 & 2.94 & 158.34 & 6.27 \\
 Trees & 1461.64 & 57.85 & 993.27 & 39.31 & 590.35 & 23.37 & 510.92 & 20.22 \\
 Water & 13.37 & 0.53 & 9.62 & 0.38 & 11.06 & 0.44 & 11.75 & 0.47 \\ \midrule
 \textbf{All} &\textbf{ 2525.88} & \textbf{99.98} & \textbf{2526.07} & \textbf{99.98} & \textbf{2255.87} & \textbf{89.29} & \textbf{2525.73} & \textbf{99.97} \\
\bottomrule
\end{tabular}
}

\end{table*}

\subsection{Impact}\label{subsec:impact}

 \paragraph*{Methodological Impact} 
The lower accuracies of global LULC maps, compared to the map produced by our local model, emphasize the importance of developing local models for more accurate LULC maps. The observed inconsistencies among these global maps underscore their limitations in African contexts, which pose adverse impacts on policy formulation and decision-making when these less accurate maps are used. Additionally, our modeling framework, comprising teacher and student models, underscores the need to effectively use diverse and growing data sources for LULC mapping. Knowledge transfer from the teacher model to the Sentinel-based student model achieves a more accurate LULC map, with the potential to scale due to the freely available nature of Sentinel-2 imagery.

  \paragraph*{Environmental and Agricultural Impact} Given the low resources available for decision makers in governmental organizations in SSA, including Murang'a county of Kenya, LULC maps provide basic insights by characterizing land cover types, enabling data-driven interventions and policy designs. Agriculture is a critical sector for most economies in SSA and hence the increased need for data-driven insights to improve it. However, the sector faces multiple challenges, including climate change resulting in rising food insecurity. LULC maps help to achieve most of the SDGs, particularly SDG 2: Zero Hunger, by enhancing food security, e.g., through efficient land use, automated crop mapping, and monitoring. Additionally, LULC maps support the compliance process to other challenges, such as the European Union's Anti-deforestation Law~\cite{Heldt2024}, by analyzing longitudinal changes of croplands.

 \paragraph*{Cross Collaborations among Diverse Organizations} This work involved a close collaboration of diverse teams of domain experts from industry, academia, and government organizations. 
 Our local partner, The Kenya Space Agency, selected Murang'a county for the pilot study. Such a collaborative effort enhances the trustworthiness of the developed product and increases the likelihood of its deployment for practical impacts. We involved end users, including the Murang'a county Government, throughout the development process. The preliminary version of our solution is deployed at the Kenya Space Agency~\cite{ksadashboard}. Partnering organizations are also using it for downstream tasks of crop type mapping, monitoring, and yield estimation.

\subsection{Limitations}\label{subsec:limitations}

\paragraph*{Data} As any machine learning framework that takes data as its main input, our framework is clearly dependent on the quality of data being used to train our models. The high-resolution Maxar imagery exhibits a heavy presence of clouds, which affects the annotation effort and minimizes the size of imagery that can be used for model training. There are still quality gaps for some of the label examples, collected from manual annotations by domain experts, partly due to the intrinsic similarity of a few LULC classes, e.g., Grasslands vs. Shrub \& Scrub. It is partly due to such annotation quality that we discarded the \textit{Flooded Vegetation} class from our analysis. Thus, more quality assurance measures could be put in place to alleviate the problem. 

\paragraph*{Methodological} Our methodology did not use the temporal information in our analysis, which is important for understanding croplands as their appearance changes across different seasons. For example, a cropland may appear as Bare Ground before planting/seeding and as Grassland or Shrubland early after the seeding season. Thus, the work could be extended to include longitudinal changes. 

\paragraph*{Evaluation} While we adopt different evaluation sets, such as Whole, Test, and External sets, and several metrics to evaluate the performance of our LULC maps, care must be taken as all the quantitative metrics are derived from a small set of manually annotated labels. These labels are sparse and only constitute a smaller percentage of the whole imagery. Further collections of label examples, from multiple annotators and the use of other existing layers, could be made to increase the size of the validation set. Thus, a more exhaustive evaluation step, including on-ground verification, is necessary for more confident validation.

\section{Conclusion and Future Work}

Food security remains a significant challenge, particularly in the Global South, including Sub-Saharan Africa, partly due to adverse climate impacts and population growth. Earth observation technologies provide a promising opportunity to improve food security by generating diverse insights using increasingly available resources, such as Sentinel images, and deep learning models. Land-use and Land-cover (LULC) maps are instrumental in resource management and environmental monitoring by characterizing key land cover types. However, existing global LULC maps were reported to be lower in accuracy and inconsistent when validated in Africa. The alarming trend of food insecurity in Africa necessitates the need for accurate LULC maps to support the agriculture sector through informed decisions for crop mapping, monitoring, and yield estimation. 

In this work, we proposed a
data-centric framework to build a local LULC mapping model with a setup of teacher and student models. We used Murang'a county in Kenya as our area of interest (AOI). 
Our framework facilitated the efficient utilization of satellite images with varying scales of resolution and availability. We used Maxar images, with $0.331$ $\mathtt{m/pixel}$ resolution, to train the teacher model and Sentinel-2 images, with $10$ $\mathtt{m/pixel}$ resolution, to train the student model. While the availability of Maxar images is tasked, expensive and limited to only the portion of our AOI, the Sentinel-2 images are freely accessible. Thus, our framework enables effective utilization of diverse data sources and build a more accurate LULC map, when 
we 
compared it with existing global maps: GDW~\cite{brown2022dynamic}, ESA~\cite{zanaga2022esa}, and ESRI~\cite{karra2021global}. 
Additionally, we observed that existing global maps not only exhibited lower accuracy but also showed inconsistencies with low agreement among themselves.
Future work includes scaling the framework to generate a LULC map for the entire country. We also plan to use temporal information to understand longitudinal changes of LULC. Temporal LULC maps can also aid in addressing compliance requirements with the recent EU Anti-deforestation Law, which poses additional challenges for small-scale farmers and producers in Africa and beyond. 

\bibliographystyle{unsrt}  
\bibliography{references}

\end{document}